# Can Q-learning solve Multi Armed Bantids?

Refael Vivanti

## Abstract

When a reinforcement learning (RL) method has to decide between several optional policies by solely looking at the recieved reward, it has to implicitly optimize a Multi-Armed-Bandit (MAB) problem. This arises the question - are currecnt RL algorithms capable of solving MAB problems? We claim that the suprising answer is no. In our experiments we show that in some situations they fail to solve a basic MAB problem, and in many common situations they have a hard time: They suffer from regression in results during training, sensitivity to initialization and high sample complexity.

We claim that this stems from variance differences between policies, which causes two problems: The first problem is the "Boring Policy Trap" where each policy have a different implicit exploration depends on its rewards variance, and leaving a boring, or low variance, policy is less likely due to its low implicit exploration. The second problem is the "Manipulative Consultant" problem, where value-estimation functions used in deep RL algorithms such as DQN or deep Actor Critic methods, maximize estimation precision rather than mean rewards, and have a better loss in low-variance policies, which cause the network to converge to a sub-optimal policy.

Cognitive experiments on humans showed that noised reward signals may paradoxically improve performance. We explain this using the aforementioned problems, claiming that both humans and algorithms may share similar challenges in decisionmaking. Inspired by this result, we propose the Adaptive Symmetric Reward Noising (ASRN) method, by which we mean equalizing the rewards variance accross different policies, thus avoiding the two problems without affecting the environment's mean rewards behavior. We demonstrate that the ASRN scheme can dramatically improve the results.

## 1 Introduction

Reinforcement Learning (RL) has shown much success during the last few years in achieving superhuman performance in many areas. During the training process, the agent interacts with the environment, and at each itterations it gets a state from the environment, decides on an action and gets a reward. The agent decides on a policy - an action for every possible state - which will desirabely maximize the acculative rewards in the long run. The ammount of different possible policies is often very large, and each have different properties such as mean reward and rewards variance. The agent must decide on one policy, and since the policy already includes the respond for every state, the desicion must be solely according to the rewards. This is a well-known problem in RL, which is called Multi Armed Bandit (MAB). In Multi-armed bandit problem a limited set of resources must be allocated between alternative choices in a way that maximizes their expected gain, when each choice's properties are only partially known at the time of allocation, and may become better understood by allocating resources to the choice. The name comes from imagining a gambler at a row of slot machines where each machine provides a random reward from a probability distribution specific to that machine. The objective of the gambler is to maximize the sum of rewards earned through a sequence of lever pulls. In our problem, the different possible policies are the bandits, their rewards is not random but does have an unknown stochastic distribution with unique properties for each bandit. Our task is a bit different from classic MAB task in several ways: The first is that while in classic MAB the amount of bandits is a low fixed number, here the ammount of possible policies is enourmos. In this work we will overlook this problem, and assume that the agent have to decide between only two policies. The second is that in classic MAB the bandits are distributed iid, which may not be the case in policies - some of them might be correlated. We will not address this problem here too. The third is that in classic MAB the different bandits usually have different mean reward but with similar variance. This is not the case in our task. Here, different policies have different mean reward, but also different rewards variance. We will focus on this problem, and especially on the common case where the two competing policies are an "interesting" one - with high mean and variance, and a "boring" one - with low mean and variance.

If any RL algorithm during any training process must implicitly solve also a MAB problem, the question arises whether it is capable of doing that. Since MAB tasks are considered relatively simpler than Markov Decision Process (MDP) ones, the common belief is that they should be able to solve that. In this work we try to establish that this is not so straight-forward. That there are several cases of MAB problems which classic RL algorithms such as Q learning are unable to optimize. And that there are common cases where they show brittle convergence, with phenomenons like high sample complexity and regression in results during training. We demonstrate those penomenons in our experiments.

We investigate for the reason that training a Q learning algorithm for solving a MAB task shows such a behaviour in the presence of difference in variance between the bandits. We identify two different problems: To the first problem we call the "Boring Policy Trap". Each policy have an implicit exploration rate, which is derived from its rewards variance. The implicit exploration of a policy is the probability that an agent will abandon this policy and choose another, due to an estimation that the later is a better one. If the original policy is "interesting" i.e. it has high rewards variance, the agent's estimation of this policy will be with high variance as well, and the probability of such a swap is higher, hence it has higher implicit exploration rate. For a "boring" policy with low rewards variance, the agent's estimation will have low variance too, the probability for a policy swap declines and this policy have low implicit exploration. This makes the probability of moving from an interesting policy to a boring one higher than in the opposite direction, and thus agents tend to get stuck at a boring policies for long periods, or even forever. If the boring policy has low mean reward, the algorithm may take a long time to converge to the better policy, or it will be unable to do it at all. The second problem, which we call the "Manipulative Consultant" problem, is unique to deep-learning algorithms in RL, which have a value-estimation functions, usually a convolutional neural network, such as the Q network in DQN or the Critic network in Actor Critic. During the training process, the minimized loss is in terms of estimation accuracy - it minimizes estimation error rather than maximizing mean rewards. We show that once the network is fully trained to the optimal policy but the training process continues, it's loss equals the policy variance, and thus any further improvment in the loss will be by swapping to a less optimal policy, which will outcome in regression in the mean reward, while improving the loss term.

Since the MAB process is implicit, and we do not have the explicit policies which the agent have to choose from, we cannot improve this process using state of the art MAB algorithms such as Thompson sampling. Thus we concentrate our efforts on reducing the variance difference between the policies. We introduce the Adaptive Symmetric Reward Noising (ASRN) scheme which is designed to learn the current policy variance and add noise to the future rewards in a way which will equalize the rewards variance through the policies. This is inspired by the cognitive experiments of Gureckis et al. [7] which showed that adding noise to the rewards might improve human performance on a similar task.

## 2  Related work

Many RL algorithms have drawbacks, such as a need for high sample complexity, high sensitivity to initialization and parameters, and regression in results during training. Often, the agent shows attraction to sub-optimal policies. A popular solution is to introduce exploration strategies to introduce other policies to the training agent, but those interfere with the behavior of the regular policy. Extrinsic exploration, such as epsilon-greedy methods [2], enforces different actions than those suggested by the policy, thus reduce the ooverall performance. Recent works introduce intrinsic exploration by rewarding the agent for its interesting behavior either as the sole reward system [3] or combined with the true rewards [4]. However, changing the reward system during training can produce a policy which behaves significantly differently from the desired one, maximizing interest rather than rewards. Eeducing the exploration gradually, as in [4], will eventually create a reward-maximizing agent, but as we will demonstrate below, once the exploration is gone even a fully-trained agent wil degrade to a sub optiaml policy. This forces selective training termination, trying to find the sweet-spot where exploration faded but degradation haven't started yet. It is desirable to have a non-exploration method which is capable of maximizing rewards and keep the acheived policy if it is optimal.

Interestingly, in cognitive experiments Gureckis et al. [7] found that adding noise in reward signals may paradoxically improve human performance. They explored the "Farming on Mars" cognitive task, in which participants were asked to repeatedly choose between two rewarding options, one with short-term greater reward but long-term less reward, and one vice-versa. They discovered that this task is very challenging for humans. When the researchers added four levels of normallydistributed noise to the rewards, they found that a high amount of noise reduces performance since the signal-to-noise ratio is too low, but low amount of noise improves performance. The idea of adding normally-distributed noise, or indeed any symmetrically-distributed noise, can be generalised as a non-exloration solution for artificial learning. This solution has the desirable property of not influencing the long-term reward behavior, so an agent training with this noise is fully-trained for the non-noised environment.

Noise is currently being used in several ways in RL. The common epsilon-greedy exploration scheme adds noise to the action space [2]. Plappert et al. [1] have suggested adding noise to the agent's parameters, for more consistent exploration. Inspired by [7], we suggest adding symmetric noise to the environment rewards in an adaptive way, aiming to equalizing the variance across different policies.

## 3  Boring Policy Trap

Our first observation is that in many environments, different policies have different properties such as reward mean and variance. We call relatively low-variance-low-mean policies "boring policies", in contrast to high-variance-high-mean policies which we call "interesting policies". Note that this is not the usual interpretation of these phrases in other RL literature such as [8]. In many cases, it seems the agent prefers boring policies during the training process.

Suppose there are only two policies, boring and interesting, and a Q-learning algorithm has to decide between them. When there is a high variance in the rewards of the interesting policy $r^t_i$, there is also a relatively high variance in the value-estimation of the interesting policies $Q^t_i$. In the rare case of a concentration of low rewards in a short period, $Q^t_i$ will be temporarily lower than the value-estimation of the boring policies $Q^t_b$. At this point the Q-learning algorithm will choose the boring policy in order to maximize its reward due to the currently available estimations. Now the training process enters what we call the "Boring Policy Trap". Once the boring policy is chosen, it will only sample examples from the boring policy. When this happens, getting back to an interesting policy, requires a complementary error, in which a series of bad boring examples reduces the current boring policies' reward-estimation $Q^t_b$ below the (currently low) $Q^t_i$. If $\sigma_b$ is low, the variance of $Q^t_b$ will be very low, so the complementary error will hardly occur and the policy will be boring most of the time. This is why we call it a "trap". In the special case of $\sigma_b = 0$, there is no exit from the Boring Policy Trap. This type of entry to the Boring Policy Trap can happen even to an well-trained agent, with true rewards estimation, if the training process continues.

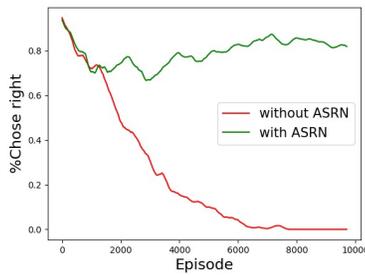
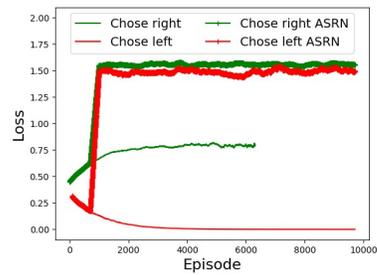

The Boring Policy Trap can also stem from the process initialization parameters. If the Q table is initialized randomly with values that are distributed i.i.d. to the environment's rewards, from among policies which may be an interesting policy and a boring policy, about half the agents will have $Q^0_b > Q^0_i$, and they will be trapped in the Boring Policy Trap for this state. The Boring Policy Trap bears resemblance to the phenomenon found by Gureckis et al. in [7], where the human participants persisted in choosing the lower-rewards option due to its low variance unless the variance was enhanced using noise.

### 3.1 The Manipulative Consultant Problem

The second problem appears when Deep RL algorithms such as DQN or Actor-Critic use a valueestimation function, or a "Consultants", for estimating the future rewards, instead of the Q table from above. It is implemented as a deep neural network which is optimized by minimizing the squared

Figure 1: Success probability during a training on the Broken-Armed-Bandit game. Without ASRN: Plain Q-learning algorithm. All agents move eventually to pull the left arm with zero reward, and never recover. With ASRN: Same, with the ASRN scheme. The symmetric noise enables a complementary error, so agents can exit the "Boring Policy Trap". time difference error loss $\delta^2$:

Figure 2: Mean loss during training. Chose Right: currently right-choosing agents. Chose left: currently left-choosing agent. Chose right ASRN: currently right-choosing agent with ASRN. Chose left ASRN: currently rightchoosing agent with ASRN. Loss differences are increasing as exploration decays. ASRN equalizes the loss.

$$\delta = Q(s_t, a_t) - (r_t + \gamma \max_a Q(S_{t+1}, a_{t+1})) \quad (1)$$

When the network is fully trained, and the environment has stochastic rewards, then:

$$\delta = \sum_{t=i}^{\infty} \gamma^{i-t}\bar{r}_i - (r_t + \gamma \sum_{t=i+1}^{\infty} \gamma^{i-t-1}\bar{r}_i) \quad (2)$$

where $\bar{r}_t$ is the mean of $r_t$

$$\delta = \sum_{t=i+1}^{\infty} \gamma^{i-t}\bar{r}_i + \gamma^0 \bar{r}_t - r_t - \sum_{t=i+1}^{\infty} \gamma^{i-t}\bar{r}_i) = (\bar{r}_t - r_t) \quad (3)$$

$$E(\delta^2) = E(r_t - \bar{r}_t)^2 = VAR(r_t) \quad (4)$$

So the consultant's loss will be lower on low-variance policies than on high-variance ones. If the low-variance policy has also lower mean reward, the loss will decrease while the agent's actual rewards will worsen.

Inspired by the cognitive experiments of Gureckis et al. [7], we propose the Adaptive Symmetric Reward Noising (ASRN) scheme, which adds a symmetric noise to the rewards, in an adaptive way, so that low variance policies will have more reward noise added to them than high variance ones. First we estimate the "interest grade" $I_t$: during the agent training, we also train several DNNs, with the same shape as the consultant network and with random initialization, to predict the current reward for each action given a state. $I_t$ is their mean absolute error compared to the actual current reward. Next, we maintain a pool of the last K interest grades, and calculate the median of $[I_{t-K}, I_t]$. Finally, to any step with interest grade less than this median, we add a symmetric noise $v$ sampled from the distribution: $v \sim N(r_t, I_t^2)$

## 4 Experimental results

The first experiment illustrates the problems described above. It is a Two-Armed-Bandit environment, where the bandits represent two competing policies with the following distributions:

$$L \sim N(\mu_l, \sigma_l^2) \text{ and } R \sim N(\mu_r, \sigma_r^2) \text{ where } \sigma_r >> \sigma_l \quad (5)$$

Since the environment has only one state, the Q-learning update expression is:

$$Q_{new}(a_t) \leftarrow (1 - \alpha)Q(a_t) + \alpha(r_t + \gamma \max_a Q(a)) \quad (6)$$

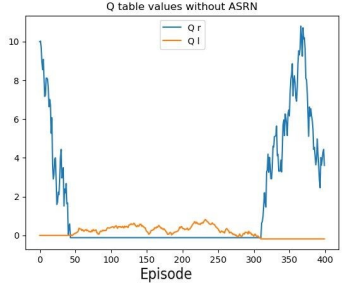
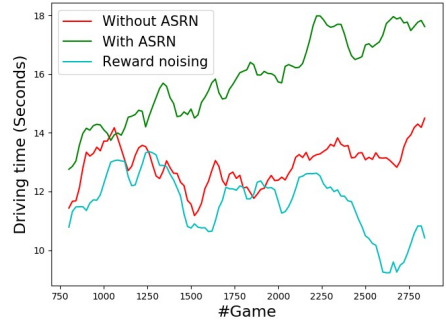

Figure 3: The parameters $Q_r^t$, $Q_l^t$ of a training agent during training. The agent enters the Boring Policy Trap at episode 40, and exits at episode 320, when the complementary error occurs.

Figure 4: Training DQN agent for self driving with AirSim. Driving time - best from last 20. Without ASRN: DQN algorithm from [12]. Reward noising: with uniform noise of $\sigma = 0.1$. With ASRN: using the ASRN scheme.

The Q-table has only two parameters, $Q_r, Q_l$, which are a cumulative-discounted-future-rewards estimation for the right and left arm, respectively. We assume $0 < \gamma < 1$, and initialize both parameters to the optimal values:

$$Q_r = \sum_{t=0}^{\infty} \gamma^t \mu_r = \frac{\mu_r}{1-\gamma} \quad \text{and} \quad Q_l = \sum_{t=0}^{\infty} \gamma^t \mu_l = \frac{\mu_l}{1-\gamma} \tag{7}$$

in order to simulate a trained policy and to show the retreat-from-success phenomenon. We implemented a decreasing epsilon-greedy exploration framework, with a low decay rate of 0.001. First we choose: $\mu_l = 0$, $\sigma_l = 0$, $\mu_r = 1$, $\sigma_r = 2.5$, $\alpha = 0.05$, $\gamma = 0.95$. We call this a "Broken-Armed Bandit", because it represents the not common situation where the left arm does nothing.

In Fig. 1 we can see the behavior of 100 players, when the graph indicates the number of players currently choosing to pull the right arm. Eventually, all of the regular players (red) decide to pull the left arm, whereas most of the players with ASRN (green) decide to pull the right arm. This demonstrates that there are some MAB tasks which Q learning is unable to optimize without help. In Fig. 2 we examine the update term of the Q-learning, $\delta_t$, which in DQN methods will be the network loss term, averaging over 100 players. We can see that without the ASRN scheme, the mean loss of players who chose to pull the right arm (thin green line) – representing choosing an interesting policy – is indeed higher than of those who chose to pull the left arm (thin red line) – a boring policy. In comparison, we can see that when we activate the ASRN scheme at step 1000, the loss is similar in both arms, and therefore the "Manipulative Consultant" problem isn't expected to take effect. To better understand the Boring Policy Trap phenomenon, we look at one training agent without the ASRN scheme and with the following parameters: $\mu_l = 0$, $\sigma_l = 0.5$, $\mu_r = 1.0$, $\sigma_r = 7$, $\alpha = 0.1$, $\gamma = 0.9$. In Fig. 3 we examine the parameters $Q_r^t$, $Q_l^t$, which are the only entries of the Q table, along the training process. We can see the policy entering the Boring Policy Trap at episode 40, when it falsely estimates that $Q_r^t < Q_l^t$, and exiting the trap at episode 320, when the complementary error occurs and the policy estimates that $Q_r^t > Q_l^t$. Exiting the trap has lower probability when the left arm has lower variance. This experiment, where the boring policy has low but not zero variance, demonstrates a more common situation, and show that Q-learning have hard time dealing with variance differences, and it spends much of the training priod estimating sub-optimal policies due to the boring policy trap, causing high sample complexity. The code used for this experiment can be found in the project's github.

## 5  Conclusions

Any RL algorithm solving an MDP problem is expected to implicitly solve a MAB problem when choosing between policies. We ask whether Q-learning and DQN algorithms are good to solve such MAB tasks. The suprising answer is that in some cases such as the broken-armed-bandit they are unable to do that, and that in other cases where there is a boring policy with relatively low rewards variance, they suffer from sensitivity to initialization and high sample complexity. We conduct experiments to demonstrate those problems. We investigate what causes those problems and find two of them: The first is the "Boring Policy Trap" problem, where an agent that decides to do boring actions has low probability of changing its mind because of the low variance. The second is the "Manipulative Consultant" problem, where value-estimation functions used in deep RL algorithms such as DQN or deep Actor Critic methods, maximize estimation precision rather than mean rewards, and have a better loss in low-variance policies, which cause the network to converge to a sub-optimal policy.

Inspired by the cognitive experiments in [7], where it is suggested that low reward noising can improve performance, we introduced the Adaptive Symmetric Reward Noising (ASRN) scheme, which adds a normal noise to rewards adaptively to the level of interest, so that their long-term mean value stays unchanged, while equalizing the noise across the policies. We showed that this can solve the Two-Armed Bandit environment even with severe variance differences.